\documentclass[runningheads]{llncs}

\usepackage[T1]{fontenc}
\usepackage{graphicx}
\usepackage{booktabs}
\usepackage{amsmath}
\usepackage{amssymb}
\usepackage{mathtools}
\usepackage[table]{xcolor}
\usepackage{hyperref}
\usepackage{cleveref}
\usepackage{tikz}
\usetikzlibrary{shapes,arrows,positioning,decorations.pathreplacing}
\usepackage{algorithm}
\usepackage{algorithmic}
\usepackage{listings}

\usepackage{todonotes}
\usepackage{pifont}
\usepackage{placeins}
\newcommand{\cmark}{\ding{51}}

\begin{document}

\title{MTDrive: Multi-turn Interactive Reinforcement Learning for Autonomous Driving}

\titlerunning{MTDrive: Multi-turn Interactive RL for Autonomous Driving}

\author{Xidong Li\inst{1}$^*$ \and
Mingyu Guo\inst{1}$^*$ \and
Chenchao Xu\inst{2}$^*$ \and
Bailin Li\inst{1}$^\dag$ \and
Wenjing Zhu\inst{2}$^\dag$ \and
Yangang Zou\inst{1} \and
Rui Chen\inst{2} \and
Zehuan Wang\inst{2}}

\authorrunning{X. Li et al.}

\institute{Li Auto Inc., Beijing, China \\
\email{\{lixidong,guomingyu,libailin,zouyangang\}@lixiang.com}
\and
NVIDIA, Shanghai, China \\
\email{\{chenchaox,wenjingz,charliech,zehuanw\}@nvidia.com}
\\[1ex]
$^*$Equal contribution \quad $^\dag$Corresponding author}

\maketitle

\begin{abstract}
Trajectory planning is a core task in autonomous driving, 
requiring the prediction of safe and comfortable paths across diverse scenarios.
Integrating Multi-modal Large Language Models (MLLMs) with Reinforcement Learning (RL) 
has shown promise in addressing ``long-tail'' scenarios. 
However, existing methods are constrained to single-turn reasoning, 
limiting their ability to handle complex tasks requiring iterative refinement. 
To overcome this limitation, we present \textbf{MTDrive}, a multi-turn framework 
that enables MLLMs to iteratively refine trajectories based on environmental feedback.
MTDrive introduces \textbf{Multi-Turn Group Relative Policy Optimization (mtGRPO)}, 
which mitigates reward sparsity by computing relative advantages across turns.
We further construct an interactive trajectory understanding dataset from closed-loop simulation to support multi-turn training.
Experiments on the NAVSIM benchmark demonstrate superior performance compared to existing methods and the effectiveness of our multi-turn reasoning paradigm.
Additionally, we implement system-level optimizations to reduce the data transfer overhead 
caused by high-resolution images and multi-turn sequences, achieving $2.5\times$ training throughput.
Our data, models, and code will be made available soon.

\end{abstract}

\section{Introduction}

Autonomous driving (AD) systems need to predict safe and comfortable trajectories 
across diverse scenarios, 
including rare but safety-critical ``long-tail'' situations~\cite{hu2022stp3,hu2023uniad}.
These scenarios are underrepresented in training data, 
where end-to-end approaches often struggle 
despite their success in common cases~\cite{hwang2024emma}.
To address this, recent work integrates Vision-Language Models (VLMs) 
into AD pipelines~\cite{jiang2024senna,tian2024drivevlm,xu2024drivegpt4}, 
leveraging their broad knowledge for better generalization.
However, VLMs still struggle with fine-grained spatial reasoning: 
even state-of-the-art models frequently misjudge ego-vehicle positioning, 
miss critical obstacles, or miscount lanes~\cite{marcu2024lingoqa,wang2025omnidrive}. 

Reinforcement Learning (RL) has emerged as a promising approach 
to improve VLMs for driving tasks~\cite{jiang2025alphadrive,shao2024lmdrive}. 
Beyond the single-turn setting, where the model must succeed on the first attempt, 
RL can be naturally extended to multi-turn interaction: 
the model proposes a trajectory, receives feedback on potential issues 
(e.g., collision risks or lane violations), 
and iteratively refines the trajectory. 
However, directly extending standard algorithms like PPO~\cite{schulman2017ppo} 
or GRPO~\cite{shao2024deepseekmath} to multi-turn settings 
introduces sparse reward challenges: 
rewards are typically assigned only at the final turn, 
leaving intermediate steps without direct supervision. 
As the number of turns grows, 
the model must determine which refinements actually contributed to the outcome---
a classic credit assignment problem that makes learning inefficient and unstable. 
Moreover, existing trajectory datasets lack support for multi-turn interactive refinement, 
and there is no established data curation pipeline for multi-turn training scenarios.

Beyond algorithmic challenges, 
infrastructure support for multimodal multi-turn RL in autonomous driving remains underdeveloped. 
Existing RL frameworks~\cite{mei2025areal,nvidia2024nemorl,sheng2024hybridflow,zhang2025slime} 
are designed for text-based reasoning tasks, 
lacking optimizations for vision-language driving models---where 
high-resolution images and extended multi-turn sequences 
create substantial computational and data transfer overhead.

To overcome these challenges, we present MTDrive, a comprehensive RL framework 
for multi-turn trajectory refinement in autonomous driving, 
encompassing data, algorithms, and system infrastructure. 
Our main contributions are summarized as follows:
\begin{itemize}
    \item \textbf{Interactive trajectory understanding dataset}: 
    We curate a multi-turn interactive dataset from a closed-loop driving simulator, 
    filtering scenarios based on key safety metrics 
    (collision, drivable area compliance, time-to-collision). 
    Trajectory understanding is formulated as a reasoning task, 
    with counterfactual questioning introduced to help the model identify critical obstacles. 
    We also design dedicated SFT data to activate the model's self-reflection capabilities 
    for iterative trajectory refinement.
    
    \item \textbf{Multi-turn Group Relative Policy Optimization (mtGRPO)}: 
    We propose mtGRPO, a novel RL algorithm designed to mitigate the sparse reward problem 
    in multi-turn environments. 
    Building upon GRPO, it ensures training stability 
    by maintaining token-level advantage consistency within each turn, 
    while employing turn-level advantages to distinguish contributions across turns. 
    The algorithm also incorporates a progressive reward mechanism 
    that encourages the model to improve its reasoning accuracy as turns increase.
    
    \item \textbf{Multimodal multi-turn RL training system}: 
    We build a dedicated RL post-training system for vision-language autonomous driving 
    based on the veRL framework~\cite{sheng2024hybridflow}. 
    To address data transfer bottlenecks caused by large image inputs and multi-turn interactions, 
    we implement targeted optimizations achieving $2.5\times$ throughput improvement, 
    which are also applicable to other multimodal RL tasks.
    
    \item \textbf{Comprehensive evaluation on real-world benchmarks}: 
    Evaluated on the NAVSIM benchmark~\cite{dauner2024navsim}, MTDrive achieves a PDMS of 96.2 
    when using privileged ground-truth perception inputs for planning evaluation, 
    and 91.1 under a more realistic setting that relies only on current-frame perception 
    with kinematic modeling for future prediction.
    This demonstrates the robustness of our multi-turn reasoning framework 
    across different levels of perceptual information availability.
\end{itemize}

Extensive experiments validate the effectiveness of our framework 
in handling complex, long-tail driving scenarios 
that require iterative trajectory refinement.

\section{Related Work}

\subsection{Vision-Language Models for Autonomous Driving}

VLMs have gained significant traction in AD, 
leveraging cross-modal alignment and zero-shot generalization 
for high-level decision-making and trajectory prediction. 
Existing VLM-based driving frameworks generally follow two paradigms. 
The first paradigm utilizes VLMs to generate high-level semantic instructions 
based on historical trajectories and multi-view images~\cite{tian2024drivevlm,wang2025omnidrive}.
These instructions or latent features then guide a downstream point-to-point planner, 
such as a diffusion-based planner or a detection head~\cite{li2024recogdrive}. 
The second paradigm, which our work follows, 
directly employs VLMs to output future driving trajectories~\cite{hwang2024emma,liu2025reasonplan,zhao2025sce2drivex}. 
Compared to the former, this approach is structurally more concise and flexible, 
while representing trajectories in natural language 
enhances the model's task comprehension and provides superior interpretability.

\subsection{Reinforcement Learning in Autonomous Driving}

Developments in RL have significantly reshaped 
the training paradigms of LLMs and VLMs. 
Optimization techniques such as PPO~\cite{schulman2017ppo}, 
REINFORCE-style variants~\cite{ahmad2024back}, 
and more efficient methods like ReMax~\cite{li2023remax}, REINFORCE++~\cite{reinforceplusplus}, and GRPO~\cite{shao2024deepseekmath} have shown remarkable success in aligning large models 
with human feedback or complex reasoning tasks.
In the AD domain, researchers have increasingly integrated RL 
to enhance policy robustness and generalization. 
For instance, RAD~\cite{gao2025rad} trains agents in photo-realistic 3DGS environments, 
CarPlanner~\cite{zhang2025carplanner} learns an auto-regressive policy for multimodal trajectories, 
and Alpamayo-R1~\cite{nvidia2026alpamayor1} employs RL to enhance reasoning-action consistency in VLAs. 
Notably, recent works like AlphaDrive~\cite{jiang2025alphadrive}, TrajHF~\cite{li2025trajhf}, 
and R2SE~\cite{liu2025r2se} have introduced GRPO
to further boost the generalization of driving policies.

\subsection{Interactive Multi-turn Reasoning in Reinforcement Learning}

Multi-turn interactive reasoning has emerged as a promising direction 
for enhancing decision-making in complex tasks. 
In AD, \textit{DriveAgent-R1}~\cite{zheng2025driveagentr1} demonstrates this potential 
by adaptively switching between linguistic reasoning and tool-assisted inference, 
achieving improvements over state-of-the-art (SOTA) VLMs. 
Frontier models in general AI (e.g., OpenAI's o3/o4~\cite{openai2025o3o4}, 
Kimi-Researcher~\cite{kimi2025researcher}, and RAGEN~\cite{wang2025ragen}) 
have achieved superior intelligence through multi-turn reasoning and iterative refinement. 
However, transitioning this capability to AD 
presents challenges in training stability and reward sparsity. 
To address this, we introduce \textbf{MTDrive} with \textbf{mtGRPO}, 
enabling effective multi-turn policy refinement for autonomous driving. 

\begin{figure}[t]
    \centering
    \includegraphics[width=\linewidth]{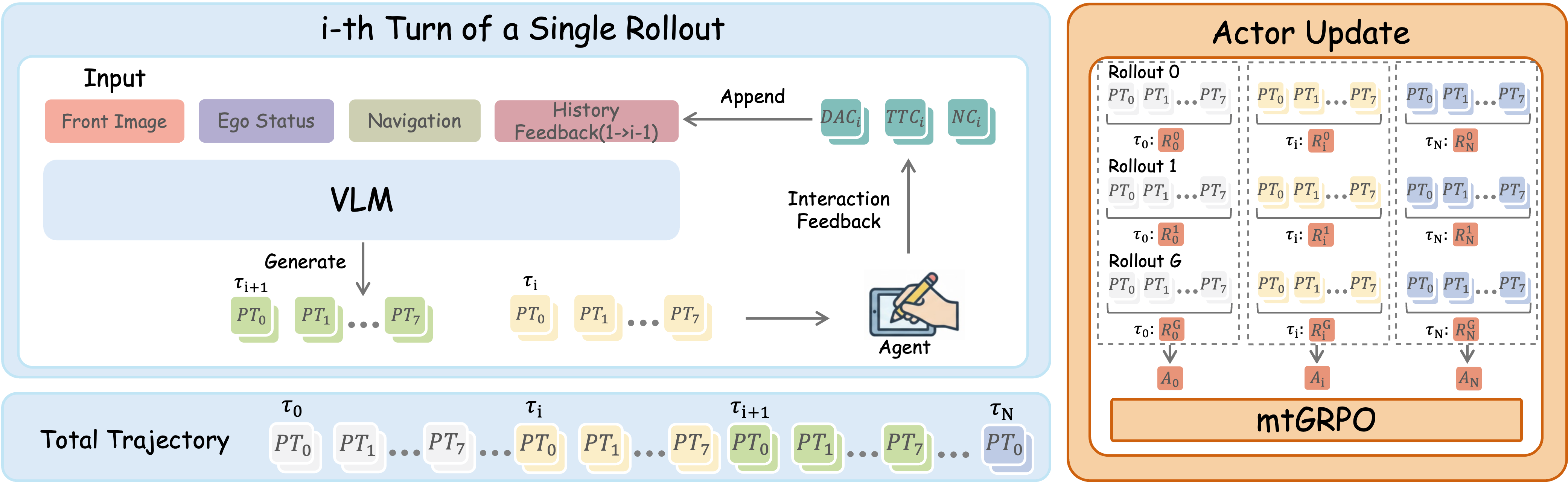} 
    \vspace{-14pt}
    \caption{The proposed MTDrive framework. 
    \textbf{Left}: Multi-turn interaction loop---at each turn $i$, the VLM takes front image, ego status, navigation, and historical feedback as input to generate trajectory $\tau_{i+1}$. The Agent evaluates the trajectory and provides per-metric feedback (e.g., collision, drivable area compliance), which is appended for the next turn. 
    \textbf{Right}: Actor update with mtGRPO---unlike standard GRPO which uses a single sequence-level reward, mtGRPO computes per-turn rewards $R_i$ and advantages $A_i$ across multiple rollouts, enabling fine-grained credit assignment for each turn's contribution.}
    \label{fig:framework}
\end{figure}

\section{Methodology}

In this section, we propose \textbf{MTDrive}, as illustrated in Fig.~\ref{fig:framework}, 
a comprehensive framework that integrates multi-turn interactive data curation, 
a novel RL algorithm (mtGRPO), and an effective multimodal RL training system.
Section~\ref{sec:pdm-agent} introduces the PDM Agent, 
which leverages collision-related metrics from the NAVSIM simulator~\cite{dauner2024navsim} 
to provide interactive feedback for trajectory refinement.
Section~\ref{sec:data} describes the data construction for both SFT and RL stages, 
including single-turn, multi-turn, and PDM understanding data.
Section~\ref{sec:rl} presents mtGRPO, 
a novel RL algorithm that computes advantages separately for each turn 
to address the sparse reward problem in multi-turn settings.
Section~\ref{sec:rl-training-system} presents a multimodal multi-turn RL training system 
built on veRL~\cite{sheng2024hybridflow}, along with 2 optimization strategies for effective training.

\subsection{PDM Agent}
\label{sec:pdm-agent}

Our multi-turn framework is designed to be agent-agnostic---it can integrate with any simulation agent that provides trajectory-level feedback. 
In this work, we instantiate it with the PDM Agent from the NAVSIM benchmark~\cite{dauner2024navsim}, 
which offers standardized metrics and enables direct comparison with existing work.

The PDM Agent is developed based on the closed-loop benchmark NAVSIM. 
Central to this framework is the PDM Score (PDMS), 
introduced in NAVSIM v1 as the primary metric for assessing driving quality. 
As formulated in Eq.~\eqref{eq:pdms}, 
the PDMS is a hybrid metric comprising 5 distinct components 
that strike a balance between safety constraints, driving comfort, and mission progress.

\begin{equation}
\label{eq:pdms}
\text{PDMS} = \underbrace{\left( \prod_{m \in \{\text{NC, DAC}\}} \text{Score}_m \right)}_{\text{penalties}} \times \underbrace{\left( \frac{\sum_{w \in \{\text{EP, TTC, C}\}} \text{Weight}_w \times \text{Score}_w}{\sum_{w \in \{\text{EP, TTC, C}\}} \text{Weight}_w} \right)}_{\text{weighted average}}
\end{equation}

After analyzing the distribution of the metrics, 
collision-related metrics---No Collisions (NC), Time-to-Collision (TTC), 
and Drivable Area Compliance (DAC)---are selected to build the PDM Agent. 
Since the PDM Agent needs to be used in both training and evaluation, 
we do not include Ego Progress (EP)
because computing EP requires access to ground-truth trajectories. 
The selected metrics are briefly described below:
\begin{itemize}
\item \textbf{NC} measures whether the autonomous vehicle (AV) collides with other traffic participants or objects.
\item \textbf{DAC} measures whether the AV consistently stays within the designated drivable area.
\item \textbf{TTC} measures whether the AV maintains a sufficient safety margin to other vehicles (typically the lead vehicle) by considering time-to-collision.
\end{itemize}

\noindent\textbf{PDM Feedback} is obtained by summarizing the violated metrics 
identified by the PDM Agent into textual prompts. 
For example, with the DAC metric, 
trajectory points outside the drivable area are extracted and described in text form. 
For TTC and NC, we similarly extract the coordinates of trajectory points 
that will result in collisions, as well as the corresponding 3D bounding boxes, 
and present these in textual format.

\subsection{Data Curation}
\label{sec:data}

\begin{figure}[t]
    \centering
    \includegraphics[width=\linewidth]{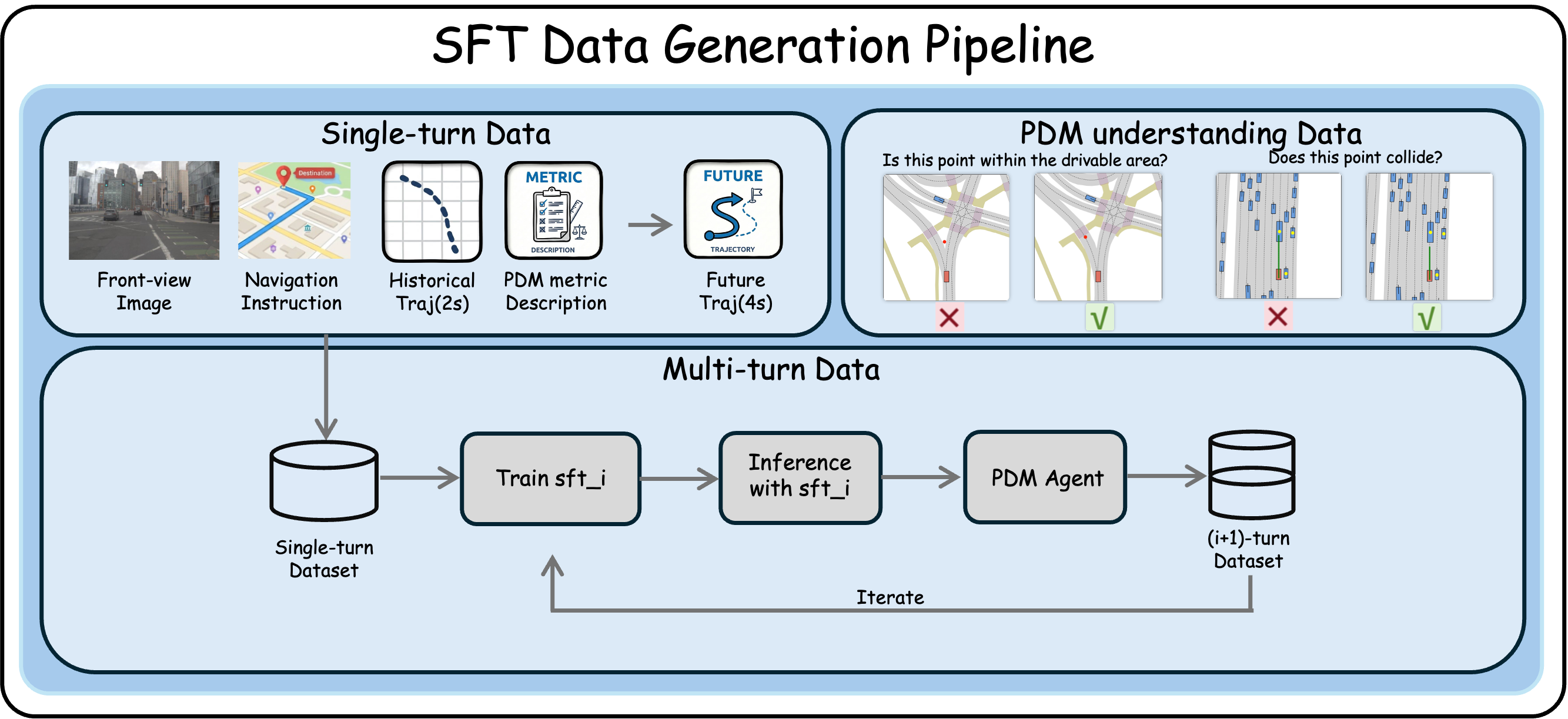} 
    \vspace{-14pt}
    \caption{Overview of the SFT Data Generation Pipeline. 
    \textbf{Top-left}: Single-turn data provides the basic trajectory generation ability which takes front-view image, navigation instruction, historical trajectory (2s), and PDM metric description as input to predict future trajectory (4s). 
    \textbf{Top-right}: PDM understanding data enables the model to interpret PDM feedback through positive/negative QA pairs.
    \textbf{Bottom}: Multi-turn data is iteratively bootstrapped from single-turn---train on $i$-turn data, run inference, obtain PDM feedback, and stack to form $(i+1)$-turn samples, enabling feedback-guided trajectory refinement.}
    \label{fig:data_pipeline}
\end{figure}

Unlike traditional multi-turn reasoning SFT (Supervised Fine-Tuning) in LLMs or VLMs, 
pretrained models for autonomous driving tasks exhibit significant limitations 
in both trajectory generation quality and instruction-following ability. 
Therefore, our SFT phase not only serves as a cold start 
to enable multi-turn trajectory generation and PDM feedback understanding, 
but also ensures sufficient trajectory quality for efficient RL sampling. 
The SFT training datasets are categorized into three types: 
single-turn data, multi-turn data, and PDM understanding data, as illustrated in Fig.~\ref{fig:data_pipeline}.

\noindent\textbf{Single-turn data.} Initially, single-turn data is included in the SFT training dataset to enable the model to acquire basic trajectory generation capabilities. 
As shown on the upper left of Fig.~\ref{fig:data_pipeline}, we adopt the trajectory dataset from RecogDrive~\cite{li2025recogdrive}, which is built on NAVSIM. 
The model receives front-view images, navigation instructions, 
2-second historical trajectory, and PDM metric descriptions, 
then predicts the 4-second future trajectory.

\noindent\textbf{Multi-turn data.} 
To activate the model's multi-turn trajectory reasoning capability, 
we add multi-turn data to the SFT training dataset. 
As shown in the second column of Fig.~\ref{fig:data_pipeline}, 
this data is constructed through an iterative bootstrap process. 

Starting from a model trained on last-turn data, 
we run inference to obtain predicted trajectories, which are then fed to the PDM agent for feedback. 
The original prompt, model prediction, and PDM feedback are concatenated 
to form the next turn's input, with the ground-truth trajectory as the target. 
This process iterates by training on $k$-turn data to generate $(k+1)$-turn samples. 
See Appendix~\ref{appendix:Multi-turn Data Curation} and ~\ref{appendix:multi-turn data} for detailed construction steps and examples.

\noindent\textbf{PDM understanding data.} 
To help the model interpret PDM feedback, 
we construct question-answering pairs for each PDM metric. 
For each PDM metric (except Comfort), we design several types of positive and negative samples. 
For example, regarding the DAC metric, 
the model is provided with a trajectory point and asked to determine whether it lies within the drivable area, as the examples in the upper right of Fig.~\ref{fig:data_pipeline} show. 
See Appendix~\ref{appendix:pdm understanding data} for detailed examples.

\noindent\textbf{RL data.} 
To improve training efficiency, 
only a subset of the NAVSIM training set is used. 
Specifically, we run evaluations on the training set using the model trained with SFT, 
and categorize the data into three types: 
(1) 2-turn data, where the PDM feedback is not empty after the first inference; 
(2) low-score data, where the PDM score after the first inference 
is below a certain threshold (0.8 in this work); 
and (3) other data. 
To construct the RL training set, we include all data from categories 1 and 2, 
and randomly sample data from category 3 to balance the distribution.

\subsection{mtGRPO}
\label{sec:rl}

In RL training for multi-turn reasoning tasks, 
directly applying GRPO leads to the sparse reward problem. 
GRPO computes a single reward for each sequence 
and uses this value for advantage calculation across the entire sequence. 
However, in multi-turn tasks, 
the performance may vary significantly across different turns within a sequence. 
For example, if the first turn performs poorly while the second turn performs well, 
but the overall sequence reward is positive, 
the poorly-performing first turn would still be rewarded, 
and the advantage of the well-performing second turn 
would be diluted by the negative influence of the first turn. 
To address this issue, we propose a novel reward design 
and the corresponding advantage calculation method for multi-turn reasoning, named mtGRPO. 

During a single RL rollout episode, 
the PDM scorer independently scores the output of each turn 
and provides multiple rewards corresponding to the number of turns in the current rollout. 
Unlike GRPO, where a single reward value is assigned to the whole sequence, 
mtGRPO assigns the reward for each turn to the corresponding tokens of that turn, 
i.e., the reward from the first turn is assigned to the tokens of the first turn, 
the second turn reward to the tokens of the second turn, and so on. 
We also incorporate a format score for each turn 
to prevent degradation of format capability during RL training. 
We provide the detailed reward calculation in Eq.~\eqref{eq:rij_formula}, 
where $p_{i, j}$ denotes the PDM score of the $j$-th turn in the $i$-th rollout, 
and $f_{i, j}$ denotes the format score of the $j$-th turn in the $i$-th rollout. 
$w_p = 0.8$ indicates the weight of the PDM score, 
and $w_f = 0.2$ indicates the weight of the format score. 

\begin{equation}
\label{eq:rij_formula}
r_{i, j} = w_p \cdot p_{i, j} + w_f \cdot f_{i, j}
\end{equation}

After the reward assignment, 
advantage estimation is performed within each turn across the rollout batch, 
following the same normalization method as GRPO. 
The detailed advantage calculation is provided in Eq.~\eqref{eq:adv_formula}, 
where $G$ denotes the rollout number 
and $\tilde{A}_{i, t}$ denotes the advantage of the $t$-th token in the $i$-th rollout. 
The indices $i$ and $t$ identify which turn $j$ the current token belongs to. 

\begin{equation}
\label{eq:adv_formula}
\tilde{A}_{i,t}
=
\frac{
r_{i,j}
-
\frac{1}{G}\sum_{i=1}^{G} r_{i,j}
}{
\mathrm{std}\!\left(\{r_{i,j}\}_{i=1}^{G}\right)
}
\end{equation}

The mtGRPO objective function is given by:
{\small
\begin{equation}
\label{eq:obj}
\begin{aligned}
\quad J_{\mathrm{mtGRPO}}(\theta)
=&\;
\mathbb{E}_{\substack{
q \sim P(Q),\;
\{o_i\}_{i=1}^{G} \sim \pi_{\theta_{\mathrm{old}}}(\cdot \mid q)
}}
\frac{1}{G}
\sum_{i=1}^{G}
\frac{1}{|o_i|}
\sum_{t=1}^{|o_i|}
\Bigg\{
\\
&\hspace{-2em}
\min \Bigg(
\frac{\pi_\theta(o_{i,t} \mid q, o_{i,<t})}{\pi_{\theta_{\mathrm{old}}}(o_{i,t} \mid q, o_{i,<t})}
\tilde{A}_{i,t},
\mathrm{clip}\!\left(
\frac{\pi_\theta(o_{i,t} \mid q, o_{i,<t})}{\pi_{\theta_{\mathrm{old}}}(o_{i,t} \mid q, o_{i,<t})},
1-\epsilon,
1+\epsilon
\right)
\tilde{A}_{i,t}
\Bigg)
\\
&\quad
-
\beta \,
\mathbb{D}_{\mathrm{KL}}
\!\left(
\pi_\theta \,\|\, \pi_{\mathit{ref}}
\right)
\Bigg\}
\end{aligned}
\end{equation}
}

The detailed procedure is illustrated in Algorithm~\ref{alg:mtgrpo}. 

\begin{algorithm}[t]
\caption{Multi-Turn Group Relative Policy Optimization (mtGRPO)}
\label{alg:mtgrpo}
\begin{algorithmic}[1]
\STATE \textbf{Input:} initial policy $\pi_\theta^{\text{init}}$, reward models $r_\phi$, task prompts $D$, hyperparameters $\epsilon, \beta, \mu$, max turn number $N$
\STATE Initialize policy: $\pi_\theta \gets \pi_\theta^{\text{init}}$
\FOR{iteration $= 1$ to $I$}
    \STATE Reference model: $\pi_\text{ref} \gets \pi_\theta$
    \FOR{step $= 1$ to $M$}
        \STATE Sample a batch $D_b$ from $D$
        \STATE Update the old policy model: $\pi_\theta^\text{old} \gets \pi_\theta$
        \STATE For each question $q \in D_b$, sample $G$ outputs $\{o_i\}_{i=1}^G \sim \pi_\theta^\text{old}(\cdot|q)$
        \STATE Compute turn rewards $r_{i j}$ for $i=1,\ldots,G$ and $j=1,\ldots,N$ using $r_\phi$
        \STATE Compute $\tilde{A}_{i,t}$ through turn-level group relative advantage estimation
        \FOR{GRPO iteration $= 1$ to $\mu$}
            \STATE Update policy $\pi_\theta$ by maximizing the GRPO objective (Eq.~\eqref{eq:obj})
        \ENDFOR
    \ENDFOR
\ENDFOR
\STATE \textbf{Output:} optimized policy $\pi_\theta$
\end{algorithmic}
\end{algorithm}

\subsection{Multimodal Multi-turn RL Training System}
\label{sec:rl-training-system}

Our training system is built on top of veRL~\cite{sheng2024hybridflow}, 
a flexible RL post-training framework that supports rollout generation, 
actor and reference model inference, and policy optimization. 
Its wide adoption and active community make it well-suited for extension to multimodal AD tasks. 
We extend veRL to support multi-turn interactions with a PDM-based environment 
for vision-language autonomous driving.

The training pipeline operates iteratively, with each step consisting of:
(1) \emph{Rollout generation}: 
the policy interacts with the PDM agent across multiple turns 
to iteratively refine trajectories and receive PDMS-based rewards;
(2) \emph{Log probability recomputation}: 
forward pass through the actor model to obtain log probabilities;
(3) \emph{Reference model inference}: 
forward pass through the reference model for KL regularization;
(4) \emph{Actor update}: 
advantage calculation and actor update via backpropagation.

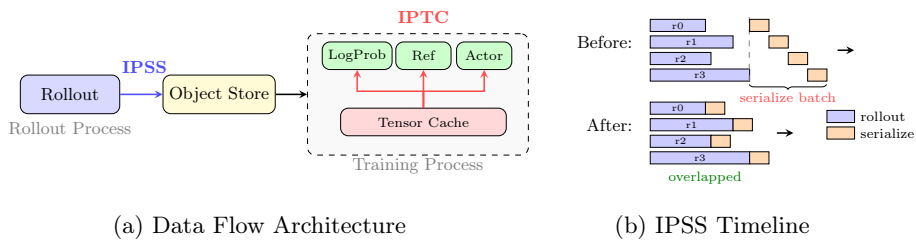
\begin{figure}[t]
    \centering
    \resizebox{\linewidth}{!}{%
    \begin{tikzpicture}[
        node distance=0.5cm,
        box/.style={draw, rounded corners, minimum height=0.7cm, align=center, font=\footnotesize},
        smallbox/.style={draw, rounded corners, minimum height=0.5cm, align=center, font=\scriptsize},
        processbox/.style={draw, rounded corners, dashed, fill=gray!5},
        arrow/.style={->, >=stealth, thick},
        rollout/.style={draw, fill=blue!20, minimum height=0.22cm, inner sep=1.5pt, font=\tiny, anchor=west},
        serialize/.style={draw, fill=orange!30, minimum height=0.22cm, inner sep=1.5pt, font=\tiny, anchor=west},
    ]
    
    \node[box, fill=blue!15, minimum width=1.8cm] (agent) at (-8.5, 0.4) {Rollout};
    \node[font=\footnotesize, text=gray] at (-8.5, -0.2) {Rollout Process};
    
    \node[box, fill=yellow!20, minimum width=1.6cm] (store) at (-5.8, 0.4) {Object Store};
    
    \node[processbox, minimum width=4.0cm, minimum height=2.2cm] (trainbox) at (-2.2, 0.4) {};
    \node[font=\footnotesize, text=gray] at (-2.2, -0.9) {Training Process};
    
    \node[smallbox, fill=green!20, minimum width=1.2cm] (oldlp) at (-3.3, 1.1) {LogProb};
    \node[smallbox, fill=green!20, minimum width=1.0cm] (ref) at (-2.1, 1.1) {Ref};
    \node[smallbox, fill=green!20, minimum width=1.0cm] (actor) at (-1.0, 1.1) {Actor};
    
    \node[smallbox, fill=red!15, minimum width=3.0cm, font=\scriptsize] (cache) at (-2.1, -0.1) {Tensor Cache};
    \draw[arrow, red!70] (cache.north) -- ++(0, 0.3) -| (oldlp.south);
    \draw[arrow, red!70] (cache.north) -- (ref.south);
    \draw[arrow, red!70] (cache.north) -- ++(0, 0.3) -| (actor.south);
    \node[font=\small, text=red!70] at (-2.1, 1.8) {\textbf{IPTC}};
    
    \draw[arrow, blue!70] (agent) -- (store);
    \node[font=\small, text=blue!70] at (-7.15, 0.9) {\textbf{IPSS}};
    \draw[arrow] (store) -- (trainbox.west);
    
    \node[anchor=east, font=\footnotesize] at (1.8, 1.35) {Before:};
    \node[rollout, minimum width=1.0cm] at (2.0, 1.65) {r0};
    \node[rollout, minimum width=1.5cm] at (2.0, 1.35) {r1};
    \node[rollout, minimum width=1.1cm] at (2.0, 1.05) {r2};
    \node[rollout, minimum width=1.8cm] at (2.0, 0.75) {r3};
    
    \draw[dashed, gray] (3.8, 1.8) -- (3.8, 0.6);
    
    \node[serialize, minimum width=0.35cm] at (3.8, 1.65) {};
    \node[serialize, minimum width=0.35cm] at (4.15, 1.35) {};
    \node[serialize, minimum width=0.35cm] at (4.5, 1.05) {};
    \node[serialize, minimum width=0.35cm] at (4.85, 0.75) {};
    \draw[decorate, decoration={brace, amplitude=2.5pt, mirror}] (3.8, 0.58) -- (5.2, 0.58);
    \node[font=\scriptsize, text=red!70] at (4.5, 0.35) {serialize batch};
    \draw[->, >=stealth, thick] (5.4, 1.2) -- (5.7, 1.2);
    
    \node[anchor=east, font=\footnotesize] at (1.8, -0.15) {After:};
    \node[rollout, minimum width=1.0cm] at (2.0, 0.15) {r0};
    \node[serialize, minimum width=0.35cm] at (3.0, 0.15) {};
    \node[rollout, minimum width=1.5cm] at (2.0, -0.15) {r1};
    \node[serialize, minimum width=0.35cm] at (3.5, -0.15) {};
    \node[rollout, minimum width=1.1cm] at (2.0, -0.45) {r2};
    \node[serialize, minimum width=0.35cm] at (3.1, -0.45) {};
    \node[rollout, minimum width=1.8cm] at (2.0, -0.75) {r3};
    \node[serialize, minimum width=0.35cm] at (3.8, -0.75) {};
    \draw[->, >=stealth, thick] (4.3, -0.3) -- (4.6, -0.3);
    \node[font=\scriptsize, text=green!50!black] at (3.0, -1.1) {overlapped};
    
    \node[rollout, minimum width=0.5cm] at (5.2, 0.0) {};
    \node[font=\scriptsize, anchor=west] at (5.7, 0.0) {rollout};
    \node[serialize, minimum width=0.5cm] at (5.2, -0.3) {};
    \node[font=\scriptsize, anchor=west] at (5.7, -0.3) {serialize};
    
    \end{tikzpicture}%
    }
    
    \vspace{2mm}
    \small
    \hspace{0.12\linewidth}(a) Data Flow Architecture \hfill (b) IPSS Timeline\hspace{0.12\linewidth}
    
    \caption{Data Transfer Optimization for Multimodal Multi-turn RL Training.
    (a) Rollout and training workers run in separate processes. 
    IPSS streams serialization during rollout; 
    IPTC enables tensor sharing among co-located modules via a shared cache.
    (b) IPSS overlaps serialization with rollout generation instead of blocking.}
    \label{fig:optimization}
\end{figure}

This iterative process introduces data transfer bottlenecks in the multimodal setting. 
In veRL's architecture, the rollout generation runs as an independent process, 
requiring all generated data including image tensors 
to be serialized and transferred through an object store. 
This leads to two sources of overhead:
(1) \emph{inter-process}: rollout results must wait until the entire batch finishes 
before being serialized and dispatched to training workers;
(2) \emph{intra-process}: even when actor and reference model are co-located, 
the controller dispatches data to each module separately, 
causing redundant deserialization of the same inputs.
We address these challenges with two complementary optimizations,
as illustrated in Fig.~\ref{fig:optimization}.

\noindent\textbf{Inter-Process Streaming Serialization (IPSS).}
In multi-turn rollout, different samples complete at varying times 
due to differences in response length. 
The naive approach waits for all rollouts to finish before serializing the entire batch, 
leaving serialization time as pure overhead. 
We observe that this serialization can be overlapped with generation: 
as each rollout completes, we immediately offload its multimodal data serialization 
to a dedicated thread pool, overlapping with the generation of remaining rollouts. 

\noindent\textbf{Intra-Process Tensor Cache (IPTC).}
When log probability recomputation, reference model inference, 
and actor training are co-located in the same process, 
the controller still dispatches data to each module separately, 
causing redundant deserialization. 
To eliminate this overhead, we implement a simple caching mechanism. 
Upon the first module's execution, IPTC caches shared inputs 
(e.g., tokenized sequences, attention masks, and visual embeddings) 
directly in GPU memory. 

Together, these optimizations reduce per-step training time 
from ${\sim}1250$s to ${\sim}490$s ($2.5\times$ speedup), 
with IPSS contributing ${\sim}1.5\times$ and IPTC 
contributing an additional ${\sim}1.7\times$ in our experiment settings 
(Section~\ref{sec:exp_setting}, Table~\ref{tab:navsim_performance}).

\section{Experiments}

\subsection{Experimental Setting}\label{sec:exp_setting}

\noindent\textbf{Supervised Finetuning.} 
We finetune Qwen2.5-VL-7B-Instruct for 6-turn dialogue with a maximum token length of 11,500. 
We use 215,000 data samples in total, 
including approximately 80,000 single-turn samples, 50,000 2-turn samples, 
5,000 multi-turn samples, and 80,000 PDM understanding samples, and train for 4 epochs. 
The learning rate is set to $4 \times 10^{-5}$, and the global batch size is 128. 
It takes about 1 day for training on a cluster of 64 A800 GPUs.

\noindent\textbf{Reinforcement Learning.} 
For reinforcement learning, we set the group size to 8, 
global batch size to 256, with a mini-batch size of 128. 
We use a constant learning rate of $1 \times 10^{-6}$, set the KL penalty coefficient to 0.01
with KL divergence computed using the K3 method. 
We use 13,000 data samples and train for 300 steps (about 6 epochs) 
on a 32-GPU cluster, which takes around 2 days.

\noindent\textbf{Evaluation.} 
During inference, we adopt the same sampling configuration 
as in the training rollout stage and set the max reasoning number to 6.

\begin{table}[!ht]
    \centering
    \caption{Performance Comparison on NAVSIM Benchmark. 
    $^\dag$ and $^\ddag$ denote supervised fine-tuning on single-turn and multi-turn data, respectively. 
    $^*$ indicates the kinematic model setting where future agent positions are predicted via constant-velocity motion models, 
    simulating practical deployment scenarios without privileged information. 
    $^{**}$ indicates the oracle setting with access to ground-truth future agent states, 
    providing an upper-bound assessment of the planning capability.}
    \label{tab:navsim_performance}
    \small
    \setlength{\tabcolsep}{6pt}
    \begin{tabular}{lccccccc}
        \toprule
        Method & NC$\uparrow$ & DAC$\uparrow$ & TTC$\uparrow$ & CF$\uparrow$ & EP$\uparrow$ & PDMS$\uparrow$ \\
        \midrule
        \multicolumn{7}{l}{\textit{Traditional End-to-End methods}} \\
        UniAD~\cite{hu2023uniad} & 97.8 & 91.9 & 92.9 & 100 & 78.8 & 83.4 \\
        TransFuser~\cite{prakash2021transfuser} & 97.7 & 92.8 & 92.8 & 100 & 84.0 & 84.0 \\
        DiffusionDrive~\cite{liao2024diffusiondrive} & 98.2 & 96.2 & 94.7 & 100 & 82.2 & 88.1 \\
        Hydra-NeXt~\cite{li2025hydranext} & 98.1 & 97.7 & 94.6 & 100 & 81.8 & 88.6 \\
        GoalFlow~\cite{xing2025goalflow} & 98.4 & 98.3 & 94.6 & 100 & 85.0 & 90.3 \\
        \midrule
        \multicolumn{7}{l}{\textit{VLM-Diffusion-based methods}} \\
        ReCogDrive~\cite{li2025recogdrive} & 97.9 & 97.3 & 94.9 & 100 & 87.3 & 90.8 \\
        ReflectDrive$^*$~\cite{li2025reflectdrive} & 97.7 & 99.3 & 93.5 & 100 & 86.9 & 91.1 \\
        ReflectDrive$^{**}$~\cite{li2025reflectdrive} & 99.7 & 99.5 & 99.1 & 99.9 & 88.9 & 94.7 \\
        \midrule
        Human~\cite{dauner2024navsim} & 100.0 & 100.0 & 100.0 & 99.9 & 87.5 & 94.8 \\
        \midrule
        QwenVL2.5-8B$^\dag$~\cite{bai2025qwen25vl} & 97.4 & 92.5 & 92.7 & 100 & 79.0 & 83.7 \\
        MTDrive$^\ddag$ (Ours) & 99.1 & 95.5 & 97.5 & 99.9 & 81.8 & 88.1 \\
        MTDrive$^*$ (Ours) & 97.5 & 98.2 & 91.8 & 99.8 & 90.6 & 91.1 \\
        MTDrive$^{**}$ (Ours) \label{tab:best} & \textbf{100.0} & 98.2 & \textbf{99.9} & 99.8 & \textbf{93.5} & \textbf{96.2} \\
        \bottomrule
    \end{tabular}
\end{table}

\subsection{Main Results}

Table~\ref{tab:navsim_performance} presents the performance of MTDrive and other models 
on the NAVSIM dataset. 
We report results under two perception settings: 
(1) \textbf{Kinematic model} ($^*$), 
where surrounding agents are assumed to move at constant velocity---a practical setting for deployment; 
(2) \textbf{GT Oracle} ($^{**}$), 
using privileged ground-truth boxes---suitable for auto-labeling applications.

Even with SFT training alone, we achieve a performance of 88.1, 
improving by 4.4 points compared to the single-turn baseline of 83.7, 
which demonstrates the strong capability of the multi-turn reasoning paradigm. 
With RL training, MTDrive achieves a PDMS of 96.2 when using ground-truth perception inputs, 
even surpassing the human driving benchmark of 94.8. 
Specifically, under the kinematic setting, MTDrive$^*$ achieves 91.1, 
demonstrating strong performance compared to traditional end-to-end methods. 
With the GT oracle, MTDrive$^{**}$ achieves 96.2, 
demonstrating the potential of multi-turn reasoning for high-quality trajectory annotation.
Notably, for the EP metric, we don't provide EP information in the multi-turn prompts 
(as supplying EP information would reveal the ground truth trajectory). 
The model attains this high EP score purely based on reward feedback during RL training.

\subsection{Ablation Study}

\begin{table}
    \vspace{-12pt}
    \centering
    \caption{Ablation Study of SFT Training}
    \label{tab:sft_data_ablation}
    \renewcommand{\arraystretch}{0.9}
    \setlength{\tabcolsep}{15pt}
    \begin{tabular}{ccccc}
        \toprule
        ID & One Stage & PDM Data & Turns & PDMS$\uparrow$\\
        \midrule
        1 & - & - & 1 & 83.7\\
        2 & & & 2 & 84.9 \\
        3 & \cmark & & 2 & 87.3 \\
        4 & \cmark & \cmark & 2 & 87.7 \\
        5 & \cmark & \cmark & 6 & \textbf{88.1} \\
        \bottomrule
    \end{tabular}
    \vspace{-12pt}
\end{table}

In this section, we present an extensive ablation study to quantify 
the contribution of each component in our method. 
The ablation studies are structured along two dimensions: SFT vs. RL, and data vs. algorithm.

For all SFT experiments, we train for a total of 6 epochs 
and select the epoch with the highest score as the final result. 
Table~\ref{tab:sft_data_ablation} shows the impact of dataset volume 
and training methods on SFT training. 
In Experiment 1, we follow RecogDrive by using only 80,000 single-turn samples 
for SFT training, serving as our baseline. 
In Experiment 2, we attempted two-stage SFT training: 
the first stage uses 80,000 single-turn samples, 
and the second stage uses 50,000 2-turn samples. 
The results show that 2-turn reasoning improves by 1.2 points 
compared to the single-turn baseline. 
In Experiment 3, we performed single-stage training 
with 80,000 single-turn samples and 50,000 2-turn samples 
and observed an increase of 3.6 points in 2-turn performance 
over single-turn performance. 
In Experiment 4, we introduced PDM understanding data, 
which led to improvements in 2-turn results with 0.4. 
Finally, in Experiment 5, we incorporated multi-turn data 
and increased the number of reasoning turns to 6, 
resulting in further gains in multi-turn PDMS scores with 88.1. 
The highest score of our best-performing 6-turn SFT model was achieved at epoch 4.

\begin{table}
    \vspace{-6pt}
    \centering
    \caption{Ablation Study of RL Strategies}
    \label{tab:rl_ablation}
    \setlength{\tabcolsep}{4pt}
    \small
    \begin{tabular}{clcccccc}
        \toprule
        ID & Method & NC$\uparrow$ & DAC$\uparrow$ & TTC$\uparrow$ & CF$\uparrow$ & EP$\uparrow$ & PDMS$\uparrow$ \\
        \midrule
        1 & GRPO (seq-level reward) & 99.9 & 96.0 & 99.7 & \textbf{99.9} & 92.0 & 94.2 \\
        2 & mtGRPO (intra-group norm) & 99.9 & \textbf{98.9} & 99.7 & 99.6 & 90.6 & 95.2 \\
        3 & mtGRPO (cross-turn norm) & \textbf{100.0} & 98.2 & \textbf{99.9} & 99.8 & \textbf{93.5} & \textbf{96.2}\\
        \bottomrule
    \end{tabular}
    \vspace{-6pt}
\end{table}

Table~\ref{tab:rl_ablation} presents the ablation of different reward 
and advantage calculation strategies. 
For the RL experiments, we use the 6-turn SFT model as the initial checkpoint 
and train for 300 steps (almost 6 epochs) on the 13,000-sample dataset 
described in Section~\ref{sec:data} and set the format weight to 0.2, as mentioned in Section~\ref{sec:rl}. 
In Experiment 1, we adopt the GRPO, 
by averaging multi-turn rewards to obtain a sequence-level reward. 
Experimental results show that, compared to the baseline SFT model, 
the PDMS score increased by 6.1 points. 
In Experiment 2, we employ mtGRPO with intra-group normalization 
for advantage calculation. 
This further increases the PDMS score by 1 point. 
Finally, when we further modify mtGRPO to use cross-turn normalization 
within the group for advantage calculation, 
the multi-turn score increases to 96.2 points.

\begin{figure}[!htb]
    \vspace{-6pt}
    \centering
    \includegraphics[width=1\textwidth,trim=2mm 2mm 2mm 2mm,clip]
    {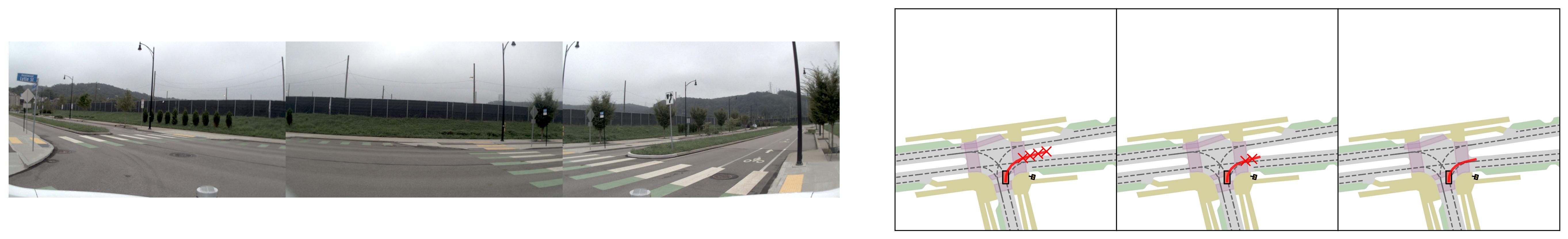}
    \vspace{-0mm}
    \includegraphics[width=1\textwidth,trim=2mm 2mm 2mm 2mm,clip]
    {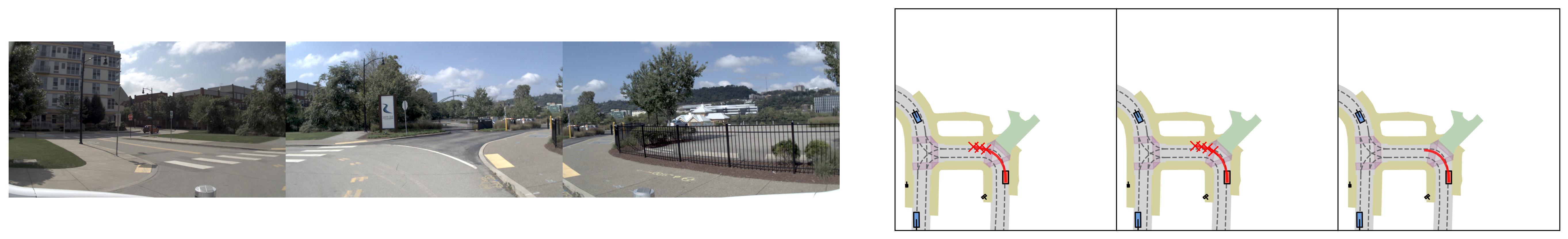}
    \vspace{-0mm}
    \includegraphics[width=1\textwidth,trim=2mm 2mm 2mm 2mm,clip]
    {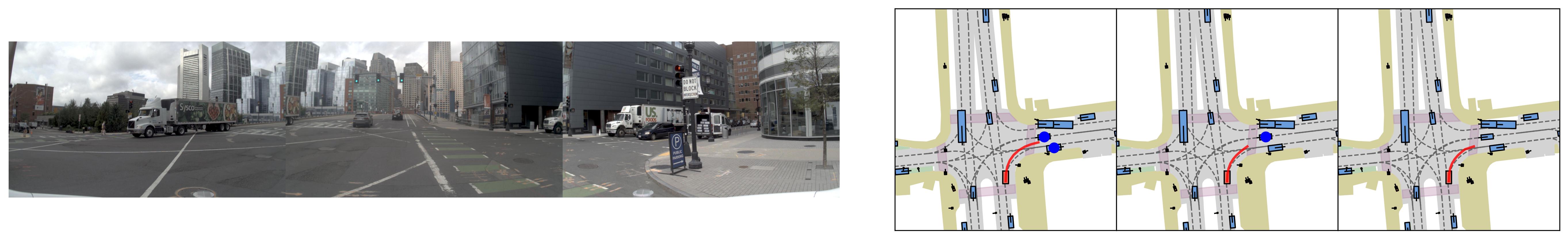}
    \vspace{-0mm}
    \includegraphics[width=1\textwidth,trim=2mm 2mm 2mm 2mm,clip]
    {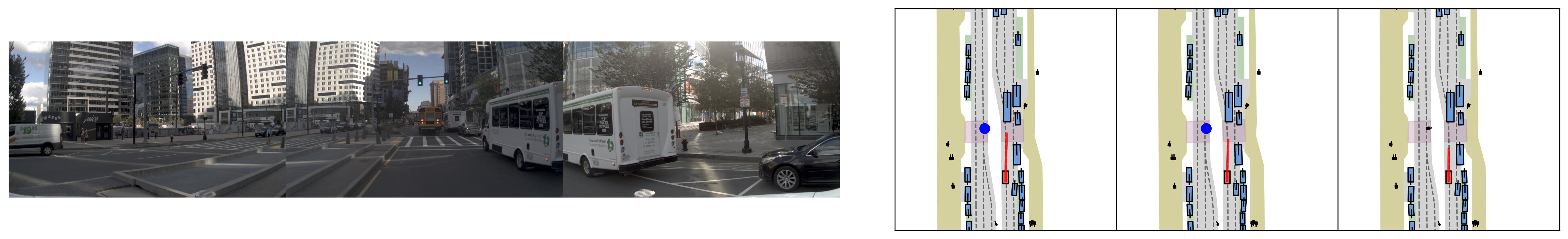}
    \vspace{-0mm}
    \includegraphics[width=1\textwidth,trim=2mm 2mm 2mm 2mm,clip]
    {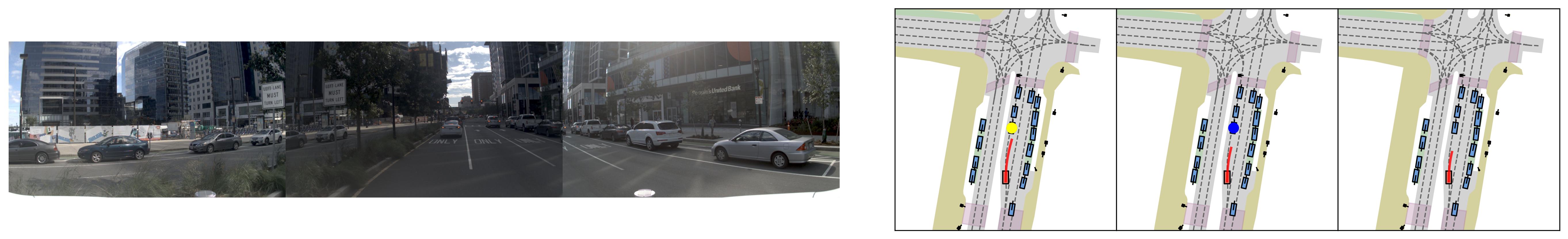}
    \vspace{-0mm}
    \includegraphics[width=1\textwidth,trim=2mm 2mm 2mm 2mm,clip]
    {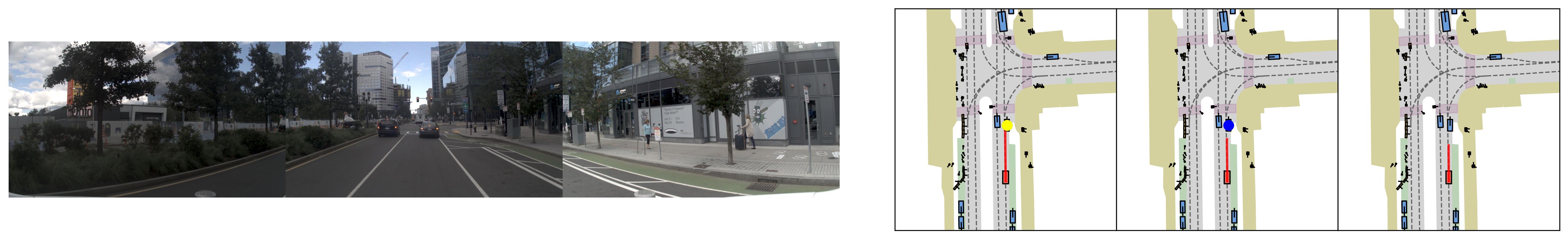}
    \vspace{-0mm}
    \includegraphics[width=1\textwidth,trim=2mm 2mm 2mm 2mm,clip]
    {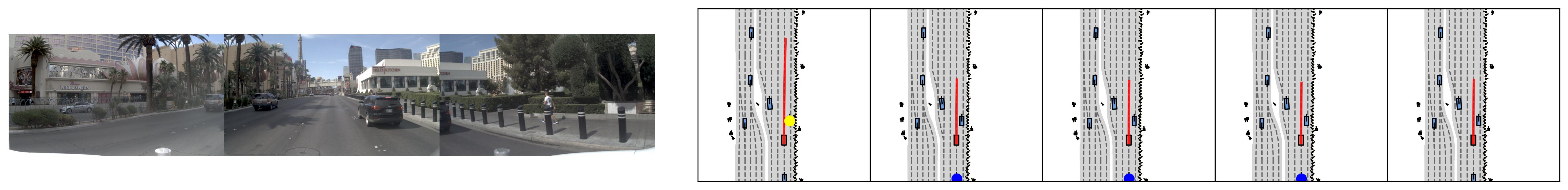}
    
    \vspace{2mm}
    \small
    \hspace{0.15\linewidth}(a) Front-view images \hfill (b) Multi-turn trajectory \hspace{0.1\linewidth}
    
    \caption{Multi-turn reasoning visualization. 
    The left figures represent images from the left, center, and right front-facing cameras, 
    while the right figures display the results of one to multi turns of reasoning. 
    In the right images, red crosses indicate trajectory points that violate the DAC metric, 
    solid blue dots indicate obstacles where potential collisions may occur according to the TTC metric, 
    and solid yellow dots indicate obstacles where collisions may occur according to the NC metric.}
    \label{fig:bev_vis}
\end{figure}

\subsection{Qualitative Results}

To further demonstrate the ability of multi-turn reasoning to improve trajectory quality, 
Fig.~\ref{fig:bev_vis} presents 7 scenarios. 
It can be observed that, during the first round of reasoning, 
the model tends to aggressively explore forward trajectories. 
Subsequently, problematic trajectory points are progressively refined based on PDM feedback, 
while non-problematic points are preserved. 

In the first 2 scenarios, the model initially generates trajectory points 
outside the drivable area which violates the DAC metric. 
In subsequent reasoning turns, the model progressively corrects the problematic trajectories 
while preserving valid points near the ego vehicle. 
In the 4th scenario, the initial trajectory generated in the first reasoning 
passes through the crosswalk, resulting in a potential collision risk with pedestrians.
After receiving PDM feedback indicating potential pedestrian activity, 
it gradually adjusts the trajectory to stop before the crosswalk to avoid the collision. 
In the 5th and 6th queueing scenarios at intersections, 
the model first triggers NC and then TTC violations, 
and ultimately maintains a safe distance from the front vehicle by the 3rd turn. 
In the last scenario, the model encounters a situation 
where the following vehicle is approaching at high speed, posing a rear-end collision risk. 
After 3 reasoning rounds, the model avoids the risk by accelerating its own speed. 
These scenarios fully showcase the rationality and superiority 
of the multi-turn reasoning approach in autonomous driving.

\section{Conclusion}

In this work, we propose MTDrive, a multi-turn VLM framework for autonomous driving. 
A multi-turn data generation pipeline is introduced to elicit the model's reflective capabilities. 
Building on this foundation, 
mtGRPO is designed to address the sparse-reward challenge in multi-turn environments. 
Finally, for multimodal multi-turn RL training, 
we implement two effective optimization techniques, IPSS and IPTC, 
to improve training throughput.

\section{Limitations and Future Work}

\noindent\textbf{Perception Dependency.} 
Our current approach is a VLM-based planner that relies on perception ground truth 
or an additional perception module to provide PDM feedback. 
A promising direction for future work is to integrate perception data into the VLM, 
enabling the model to generate PDM feedback solely based on its own reasoning. 
Such perception tasks are also expected to enhance the model's understanding of the environment.

\noindent\textbf{Auto-Labeling.} 
Since our current metrics even surpass those of human drivers, 
we believe that the multi-turn reasoning framework holds promise for trajectory annotation, 
offering a high-quality alternative to human-driver data. 
This could help address the lack of multimodal data in imitation learning for autonomous driving.

\noindent\textbf{Generalization to Other Agents.} 
While we instantiate our framework with the rule-based PDM Agent, 
the multi-turn interaction paradigm can work with any agent that provides trajectory-level feedback. 
Future work could explore learned reward agents or 
adapt the framework to agents in other simulation environments 
(e.g., CARLA~\cite{dosovitskiy2017carla}, Waymax~\cite{gulino2024waymax}, AlpaSim~\cite{nvidia2025alpasim}).

\FloatBarrier
\bibliographystyle{splncs04}
\bibliography{ref}

\newpage
\begin{center}
    {\LARGE\bfseries Appendix}
\end{center}
\vspace{1cm}

\appendix 
\section{Prompt Example}

\subsection{multi-turn data}
\label{appendix:multi-turn data}

\textbf{Question}: <image>\textbackslash nAs an autonomous driving system, predict the vehicle's trajectory based on:\textbackslash n1. Visual perception from front camera view\textbackslash n2. Historical motion context (last 4 timesteps):   - t-3: (-9.44, +0.02, -0.02)    - t-2: (-6.21, -0.02, 0.0)    - t-1: (-3.05, -0.03, 0.0)    - t-0: (0.0, 0.0, 0.0)\textbackslash n3. Active navigation command: [GO STRAIGHT]\textbackslash nOutput requirements:\textbackslash n- Predict 8 future trajectory points\textbackslash n- Each point format: (x:float, y:float, heading:float)\textbackslash n- Use [PT, ...] to encapsulate the trajectory\textbackslash n- Maintain numerical precision to 2 decimal places<|im\_end|>\textbackslash n<|im\_start|>assistant\textbackslash nHere is the planning trajectory [PT, (+3.1, 0.0, 0.0), (+6.21, 0.0, 0.0), (+9.31, 0.0, 0.0), (+12.42, 0.0, 0.0), (+15.52, 0.0, 0.0), (+18.62, 0.0, 0.0), (+21.73, 0.0, 0.0), (+24.83, 0.0, 0.0)].<|im\_end|><|im\_start|>user\textbackslash nThis was your previous trajectory prediction. Analyze this trajectory data and propose a revised prediction. Consider the following identified issues:\textbackslash n1. The TTC metric reveals that certain trajectory points fail to maintain a safe distance from surrounding objects. The format for describing these objects is: (x, y, z, l, w, h, heading, name). Where x, y, z are the center coordinates of the object in ego-coordinate system, l, w, h are length, width, height of the bounding box, heading is the object heading, and name is the object class name. The specific trajectory points and their corresponding objects are as follows:\textbackslash nTrajectory Point 1: (+21.73, 0.0, 0.0) with Objects: (34.14, -0.85, 1.59, 10.26, 2.95, 4.39, -0.02, vehicle).\textbackslash nBased on these issues, provide an improved trajectory prediction.\\

\noindent\textbf{Answer}: Here is the planning trajectory [PT, (+2.97, +0.02, 0.0), (+5.82, +0.02, 0.0), (+8.54, 0.0, -0.03), (+11.17, -0.07, -0.05), (+13.73, -0.22, -0.09), (+16.21, -0.56, -0.16), (+18.62, -1.10, -0.24), (+20.94, -1.77, -0.28)].

\subsection{pdm understanding data}
\label{appendix:pdm understanding data}

\textbf{Question}: <image>\textbackslash nYou are a vehicle trajectory prediction model for an autonomous driving system. One of your key metrics is Drivable Area Compliance (DAC), which assesses whether trajectory points remain within the drivable area. Given a trajectory prediction point and a front-view image, where trajectory point is formatted as (x: float, y: float, heading: float), your task is to identify if the given trajectory point stays within the drivable area.\textbackslash nTrajectory point is (+24.87, 0.0, 0.0).\textbackslash nOutput Requirements:\textbackslash nFormat: Yes or No\textbackslash nMaintain numerical precision to two decimal places.\\

\noindent\textbf{Answer}: No.

\section{Multi-turn Data Curation}
\label{appendix:Multi-turn Data Curation}

We first train Qwen2.5VL on single-turn data, 
this model is referred to as \texttt{Qwen2.5VL\_sft1}, 
to obtain a model with basic trajectory generation abilities. 
We then use \texttt{Qwen2.5VL\_sft1} to perform inference on the single-turn dataset, 
obtaining sequences of predicted trajectories. 
By feeding the generated trajectories into the PDM agent, 
we obtain the corresponding PDM feedback. 
Then we concatenate the prompts from the single-turn data, 
the model-generated trajectories, and the corresponding PDM feedback, 
to construct the prompt for the second turn data, 
using the same ground-truth from the single-turn data as the second turn's ground-truth. 
We further employ the constant velocity model provided by NAVSIM to follow the above steps and generate additional 2-turn data, because 2-turn samples from \texttt{Qwen2.5VL\_sft1} are far fewer than single-turn ones.

Then we follow the above steps with 2-turn data to get \texttt{Qwen2.5VL\_sft2} and 3-turn data. For the subsequent turn data, we did not continue the above process due to computational resource constraints. Instead, we directly use the three-turn model to construct data for subsequent turns. 
Specifically, we run the \texttt{Qwen2.5VL\_sft2} model multiple times on the same data 
to produce different 3-turn sequences, and stack these sequences together, 
which we refer to as mock multi-turn data. 
Since our mock multi-turn data is not entirely derived from real agent inference results, 
its distribution does not match real multi-turn reasoning, 
we only use a subset of it to stimulate the model's multi-turn trajectory reasoning capability. 
Now we have obtained all multi-turn data, 
including 2-turn, 3-turn, and mock multi-turn data.

\section{Additional Ablation}
\label{appendix:Additional Ablation}
We also study the impact of dataset size on RL performance. 
This ablation was performed using an earlier version of our approach; due to resource constraints, we are unable to rerun it with the latest version. 
Therefore, we present indicative conclusions instead of reporting precise quantitative results. 
When training with only 7,000 hard samples described in Section~\ref{sec:data}, the PDM score exhibits a minor increase within the first 30 steps, but then quickly declines. 
Evaluation on the 7,000 training set reveals clear signs of overfitting. 
Then we increase the dataset size to 13,000, 26,000, and 39,000 samples; the overall trend remains similar to that observed with 13,000 samples: convergence is slightly slower, but the performance metrics at 300 steps are very close. 
\end{document}